# Solving the Periodic Timetabling Problem using a Genetic Algorithm

**Diego Arenas**
Phd. Student, UVHC-IFSTTAR, France
**Rémy Chevrier**
Researcher, IFSTTAR, France
**Saïd Hanafi**
Professor, UVHC, France
**Joaquin Rodriguez**
Researcher, IFSTTAR, France

**SUMMARY**

In railway operations, a timetable is established to determine the departure and arrival times for the trains or other rolling stock at the different stations or relevant points inside the rail network or a subset of this network. The elaboration of this timetable is done to respond to the commercial requirements for both passenger and freight traffic, but also it must respect a set of security and capacity constraints associated with the railway network, rolling stock and legislation. Combining these requirements and constraints, as well as the important number of trains and schedules to plan, makes the preparation of a feasible timetable a complex and time-consuming process, that normally takes several months to be completed. This article addresses the problem of generating periodic timetables, which means that the involved trains operate in a recurrent pattern. For instance, the trains belonging to the same train line, depart from some station every 15 minutes or one hour. To tackle the problem, we present a constraint-based model suitable for this kind of problem. Then, we propose a genetic algorithm, allowing a rapid generation of feasible periodic timetables. Finally, two case studies are presented, the first, describing a sub-set of the Netherlands rail network, and the second a large portion of the *Nord-pas-de-Calais* regional rail network, both of them are then solved using our algorithm and the results are presented and discussed.

## 1. INTRODUCTION

A timetable defines the departure and arrival times for all trains at their scheduled stop stations or other relevant points in the rail network. The commercial requirements, i.e. the number of train lines and their frequency are defined by the train operating companies, for both passenger and freight traffic. The timetables have to respond to these commercial needs, but also, and in order to be feasible or viable, the timetable must observe a set of security and capacity constraints e.g. to respect the minimum distance (or headway time) between two trains running on the same direction on the same railway. These constraints are often called hard constraints, which means that they can not be violated under any circumstance. There is another type of constraints, the so-called soft constraints, which deal with some specific requirements for the timetable, e.g. to establish particular itinerary connections between two passenger train lines, the violation of these is allowed, however, it often leads to inefficiency, additional costs or financial loses.

Furthermore, some of the commercial requirements can enter in conflict with each other, or with capacity constraints or even with maintenance activities, these conflicts should be rapidly identified because, in most cases, negotiations need to take place between the operating companies and the infrastructure manager companies in order to solve them.

The timetabling problem represents only the first step of an even larger and more complex process, which is the yearly rail service planning. As its name suggests it, it consists of the definition of the plan, and the allocation of resources to provide the annual train services, i.e., timetables, crew schedules and rolling stock usage. The process is described in Fig. 1 and will be shortly described.

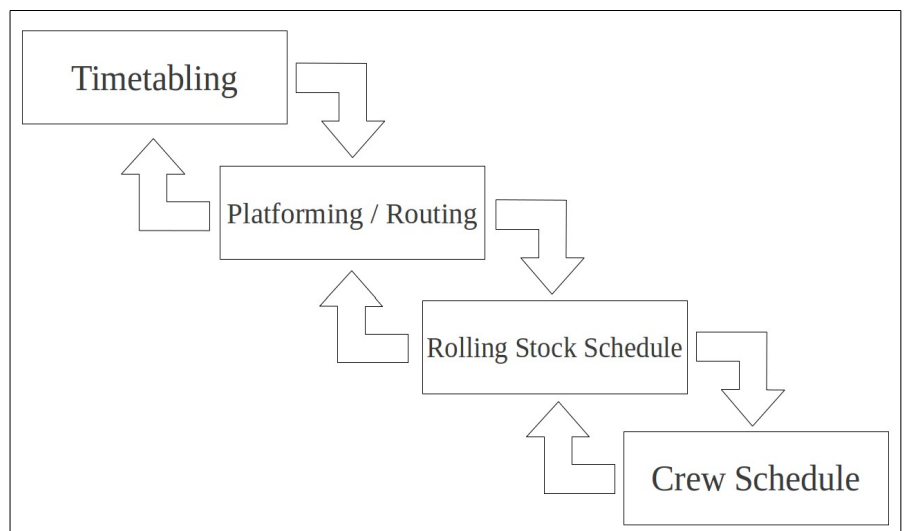

**Fig. 1 – Yearly Rail Service Planning Process**

Once a feasible timetable is obtained and the preliminary conflicts are solved, a route plan has to be designed. This is known as the routing and platforming problems, and consists of the allocation of platforms, inbound and outbound routes for the scheduled trains into the stations. At this point, some new conflicts may arise, e.g. one station has 3 trains scheduled to stop at a given time, but it only has one platform, consequently it is necessary to step back, and generate a new timetable. In some cases, many iterations are required to finally obtain a feasible timetable with its viable route plan. Ideally, the platforming and the timetabling problem should be tackled at the same time, however, real-size problems are too large and have thousands of constraints.

The other steps of the process are: rolling stock planning, which specifies among other things, what engines and carriages will be used for each scheduled train of the timetable. Finally, the crew schedule, which defines the working plan for each person (role) involved in the train operations, e.g. drivers, controllers, etc. Of course, at these steps, new conflicts may be identified, which will require to step back and change the previously generated plans. In this paper, we focus on the timetabling problem, i.e., generating a feasible timetable. Therefore the platforming and routing plans are not further discussed, and also, we assume that each station has enough platforms to accommodate the trains that will stop in it.

It is also important to note that in each of the steps previously described, an optimization process can be carried out, e.g. optimize a timetable by reducing the total number of rolling stock required, or by minimizing the overall waiting time for passengers during connections.

Traditionally, the timetabling problem is solved based on the experience and expertise of the involved railway planners. But at the early nineties it was recognized within the rail community that the application of models and techniques for supporting the solution process of these problems is truly advantageous. It not only leads to better solutions, but it also conducts to a considerable reduction of the time needed for the planning process.

There are mainly two types of timetables, periodic and non-periodic. The difference between them is the following. On the one hand, non-periodic timetables are designed based only on a set of requests for trips in the network, i.e., set the departure and arrival times as close as possible to their corresponding required departure and arrival times for a particular trip. On the other hand, periodic timetables are built in function of a cycle time T, such that a timetable pattern is created for this period where all trips are scheduled. Then, this pattern is repeated $\{kT, kT+1, kT+2, ..., (k+1)T\}$. This means that each line of passenger trains is operated in a regular recurrent way, e.g., the trains of the same line run every 30, 60 or 120 minutes (see Table 1 for an example of a periodic timetable). Periodic timetables are largely used in European countries while non-periodic timetables are often used in America and Australia.

| From | To | Departure | Arrival |
|---|---|---|---|
| Paris Nord | Lille Flandres | 8:46 | 9:48 |
| Paris Nord | Lille Europe | 9:46 | 10:45 |
| Paris Nord | Lille Flandres | 10:46 | 11:48 |
| Paris Nord | Lille Europe | 11:46 | 12:45 |

**Table 1 – Periodic Timetable for two TGV (High Speed Train) lines connecting the cities of Paris and Lille (France).**

One advantage of a periodic timetable is that passengers can easily remember and adequate to their routine the departure time of their train at their station. However, a drawback is that these kinds of timetables are not very flexible, and it is difficult to offer a large number of direct connections. Another disadvantage, is the fact that a completely periodic timetable may be rather inefficient (Borndörfer and Liebchen, 2007), trains might have to be operated even at some specific times of a day with only a small number of passengers. Therefore, in practice there are usually exceptions to the completely periodic timetable. For example, additional trains are added during rush hours, and some other trains are canceled during late evening hours (Kroon et al. , 2007). In this paper we discuss only the generation of periodic timetables.

The rest of this paper follows the next structure. Section 2 reviews some popular approaches that address this problem. Section 3 presents the formulation used in our approach. Section 4 describes how we use a genetic algorithm to solve the train periodic timetable generation problem. Section 5 presents the case studies and discusses the obtained results using our technique. Finally, Section 6 concludes the paper and presents some perspectives for further works.

## 2. LITERATURE REVIEW

Several authors have developed different techniques to address scheduling problems and specifically the train timetabling problem, a review of these can be found on (Lusby et al. 2011) and (Cacchiani and Toth 2012). In this section we will briefly discuss some of the most relevant works that deal with this topic.

Introduced by Serafini and Ukovich (1989), the Periodic Event Scheduling Problem (PESP) is a framework for scheduling periodic activities. The PESP is proven to be an exceptionally rich model that can be used in many applications, e.g., airline scheduling, traffic light scheduling and of course, train scheduling, among others. Odijk (1996) proposes the PCG (PESP Cut Generation) algorithm to generate timetables for train operations using a mathematical model consisting of periodic time window constraints as input. His model is based on the definition of constraints applied on a mathematical formulation to solve the problem.

Kroon et al. (2007a) separate the problem in two parts: the generation of the departure/arrival times and the selection of the routes through the stations. The first part describes a mathematical formulation based on the PESP to solve the scheduling problem, the authors define a set of different types of constraints and then use them as an input to the PESP solver, obtaining a timetable that will be used as an input to the second part: the routing. Additionally, Kroon et al. (2007b) presents an optimization model used to minimize the average delays of trains.

Liebchen's (2008) work is also based on the PESP, however two optimization variants are presented and discussed: The Max-T-Pesp, and a heuristic method called Cut-Heuristic. Then, these two techniques are applied to solve an optimization problem on a real scenario: the subway system of Berlin.

Caprara et al. (2002) concentrate on the problem of a single, one-way track linking two major stations, with a number of intermediate stations in between. They use a directed multigraph in which nodes correspond to departures and arrivals at certain station at a given time instant. Then, they use this formulation to derive an integer linear programming model which is relaxed in a Lagrangian way. The objective is to maximize the sum of the "profits" of the scheduled trains, the profit achieved for each train depends on: the "shift" defined as the absolute difference between the departure times from a given station in the ideal and actual timetables; and on the "stretch" defined as the non-negative difference between the running times in the actual and ideal timetables. Although their model is not applied to create a periodic timetable, it is highly flexible and it can be extended to the PESP model and then used to generate periodic timetables.

Other relevant techniques based on the PESP are Kroon and Peeters (2003), Caimi et al. (2007) and Liebchen and Peeters (2002). Moreover, some authors use evolutionary techniques such as genetic algorithms to completely or partially address the problem. As discussed by Liebchen et al. (2008), Genetic Algorithms are proven to be a solid method to deal with the periodic timetabling problem, and their behavior is more stable than other techniques when used both in small and big instances of the problem.

As an instance of partial use of genetic algorithms, Semet and Schoenauer (2005) focus on the reconstruction of a timetable following a small perturbation. They try to minimize the total accumulated delay by adapting the departure and arrival times for each train and allocation of resources. They use a permutation-based evolutionary algorithm to gradually reconstruct the schedule. The railway network can be seen as a graph where nodes are stations or switches and where interconnecting edges eventually hold several tracks for trains to use. Their objective function consists in minimizing the total accumulated delay (for all trains at all nodes). The algorithm is ah hybrid algorithm, in the sense that it combines an evolutionary engine with a mathematical programming tool, the ILOG CPLEX. The evolutionary part of the algorithm is used to quickly obtain a good but suboptimal solution, which is fed to CPLEX as an initial solution which will search for the global optimum.

An example of complete use of genetic algorithms to address the problem has been proposed by Kwan and Mistry (2003), who use a co-evolutionary algorithm for the automatic generation of train timetables. The objectives are: first, to allocate as many capacity requests as possible, and second: to discover the conflicts that have to be solved by the train operating companies. They assign some penalty weights to each type of constraint violation, thus the objective function is to minimize the weighted sum of violations. They use three types of populations which are evolved one after another: "departing times", "running times" and "resources options". The combination of three individuals, i.e., one of each population, result on a timetable. Within each step of the evolution, a timetable is generated and evaluated. Once the termination condition is achieved, the algorithm is stopped and the best timetable is given as result.

## 3. PROBLEM FORMULATION

In this section, we give the model for our solution. We define the data and the variables, then the objective function, and finally, we describe the constraints that our model takes into account.

### 3.1. Definition of data and variables
Table 2 and Table 3 give a definition of the main symbols used in our model for data and variables respectively.

| Symbol | Definition |
| --- | --- |
| $T$ | Period for one cycle. |
| $i, j$ | Trains. |
| $s, t$ | Consecutive stations. |
| $Lx, Ux$ | Lower and Upper value bounds for constraint type $x$. |
| $x$ | *Type of constraint, can acquire one of the following values: r,h,s,c. Corresponding respectively to: Running-time, Headway, Single-track and Connection.* |

**Table 2 – Symbols used to represent data in the model.**

| Symbol | Definition |
| --- | --- |
| $D_{i,s}$ | Departure time of train *i* from station *s*. |
| $A_{i,s}$ | Arrival time of train *i* at station *s*. |
| $e, e'$ | Represents events, an event is either the departure or arrival of a train from a station. |
| $Q_{e,e'}$ | Binary value. If the events *e* and *e'* take place on the same period *T* the value of this variable is 0, otherwise, is 1. |

**Table 3 – Symbols used to represent variables in the model.**

### 3.2. Objective Function

The objective function is defined as the minimization of the weighted sum of constraint violations. The weight for each type of constraint violation depends on the importance of the constraint, e.g., violating a headway constraint is much more serious than violating a connection constraint:

$$Minimize[(W_r \cdot V_r)+(W_d \cdot V_d)+(W_h \cdot V_h)+(W_s \cdot V_s)+(W_c \cdot Vc)] \quad (1)$$

$$Minimize \sum (W_x \cdot V_x) \quad (2)$$

where $W_x$ is the weight value corresponding to a violation of a constraint type $x$, and $V_x$ represents the constraint violation count of type $x$.

### 3.3. Constraints

**Running Time**

This constraint will ensure that the travel time for some train $i$ departing from station $s$ and arriving at station $t$, is between the specified limits $Lr$ and $Ur$ (Lower Running Time and Upper Running Time, respectively) previously calculated. It is defined as follows:

$$Lr_{i,s,t} \leq A_{i,t} - D_{i,s} + T \cdot Q_{arr_{i,t}, dep_{i,s}} \leq Ur_{i,s,t} \quad (3)$$

where $Lr_{i,s,t}$ and $Ur_{i,s,t}$ are determined from the details of the infrastructure between the stations $s$ and $t$, the safety rules and the running and braking time of the involved rolling stock.

**Dwelling Time**

This constraint will guarantee that the stopping time of train $i$ at station $s$ is between the $Ld$ and $Ud$ limits. (Lower Dwelling Time and Upper Dwelling Time, respectively):

$$Ld_{i,s} \leq D_{i,s} - A_{i,s} + T \cdot Q_{dep_{i,s}, arr_{i,s}} \leq Ud_{i,s} \quad (4)$$

where $Ld_{i,s}$ and $Ud_{i,s}$ are predefined values guaranteeing that there is enough time for passenger to board and alight the train $i$. Also, if the train $i$ has a connection at station $s$, this dwelling time should be large enough to accommodate the connection.

**Headway Time**

Two trains $i,j$ departing from station $s$, must have a headway time between them limited by $Lh$ and $Uh$ (Lower Headway Time and Upper Headway Time, respectively):

$$Lh_{i,j,s} \leq D_{i,s} - D_{j,s} + T \cdot Q_{dep_{i,s}, dep_{j,s}} \leq Uh_{i,j,s} \quad (5)$$

where $Lh_{i,j,s}$ and $Uh_{i,j,s}$ are calculated values based on the train types, their running times, and their proper basic headway time. The basic headway time is the minimum time that a train following another must observe. This constraint prevents the generation of schedules in which some train overtake another on the railway. The lower bound $Lh_{i,j,s}$ is defined as the difference between the minimum running times for trains $i$ and $j$ departing from station $s$ to the next station $t$, plus the maximum basic headway time of train $i$. The upper bound $Uh_{i,j,s}$ is the difference between the cycle $T$ and the basic headway time for train $j$.

**Single Track - Opposite Trains**

Similar to the headway constraint, it is used when there is only one track connecting two stations and there is a train *i* going from station *s* to *t*, and another train *j*, departing from *t* and arriving at *s* . The time between the departure of train *i* and the arrival of train *j* from/at station *s* must be within limits *Ls* and *Us*, this constraint is defined as follows:

$$Ls_{i,j,s} \leq D_{i,s} - A_{j,s} + T \cdot Q_{dep_{i,s}, arr_{j,s}} \leq Us_{i,j,s} \tag{6}$$

where $Ls_{i,j,s}$ is two times the minimum running time from the station *s* to the station *t* plus the basic headway time for the train *i*. The value of $Us_{i,j,s}$ is the difference between cycle time *T* and the minimum headway for the train *j*.

**Connections**

This type of constraint allows the connections between trains and the passenger transfers. The time between the arrival of train *i* and the departure of train *j* must be within the limits of *Lc* and *Uc* (Lower Connection Time and Upper connection time, respectively):

$$Lc_{i,j,s} \leq D_{j,s} - A_{i,s} + T \cdot Q_{dep_{j,s}, arr_{i,s}} \leq Uc_{i,j,s} \tag{7}$$

where $Lc_{i,j,s}$ and $Uc_{i,j,s}$ are defined as the estimated necessary time for a passenger arriving on train *i* at station *s*, to alight, find the connection platform and board the train *j*. Larger stations normally require longer connections times.

**4. A GENETIC ALGORITHM BASED SOLVING METHOD**

Our proposal to address this problem is to develop a genetic algorithm inspired by the models of Kwan and Mistry (2003) and Semet and Schoenauer (2005). A comprehensive description about genetic algorithms can be found in Goldberg (1989) and Holland (1992).

Our model takes into account the objective function (2) and constraints (3) to (7) described in Section 3 and it performs the generation of periodic train timetables. The two main components are detailed below.

**4.1. Data Representation**

To represent a timetable, we use one vector of integer values. The length of the vector is equal to the sum of two times the trips count for all trains, i.e. $\sum_{i \in Trains} 2 \times trips\, count(i)$ .

The vector is divided into small sections, each section corresponding to a train, inside one section, the first element represents the departure time of the train at its first station, the possible values accepted are defined as $[0...(T-1)]$ . The rest of the elements in the section represent the running and dwelling times for all trips and scheduled stops of the train. Figure 2 gives an example of a section. Each element representing a running time, will only allow values between those defined by its respective running constraint, the same is applied to the elements representing dwelling times. This will ensure that every timetable generated will automatically respect all running and dwelling time constraints.

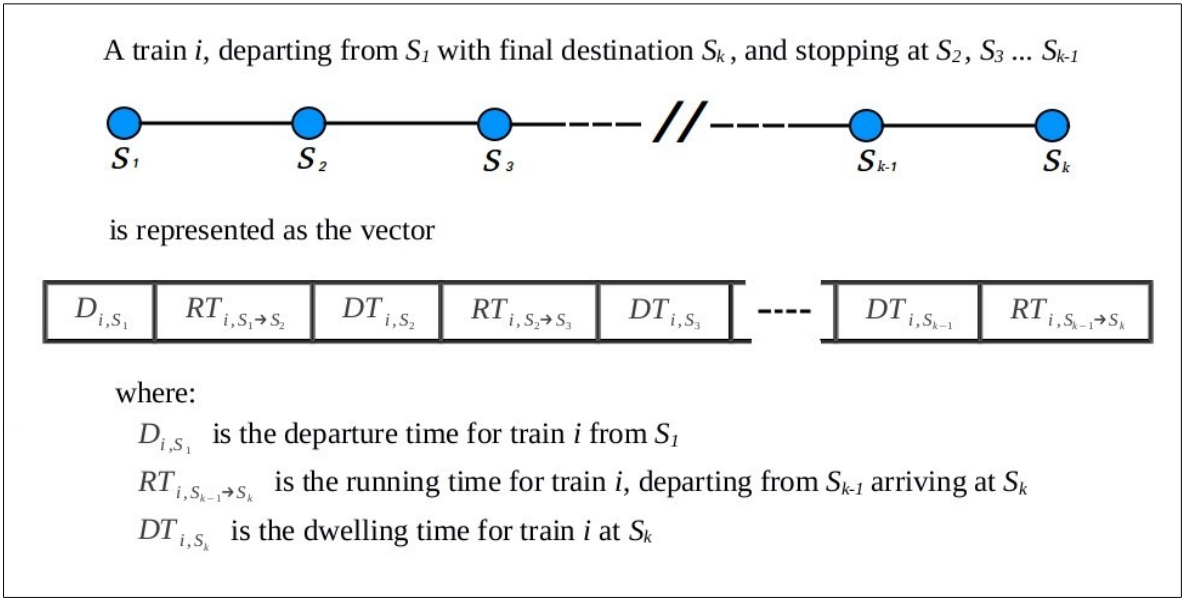

**Fig. 2 – Data representation, a train as a vector.**

 4.2. The algorithm

The main algorithm is described in Algorithm 1. The essential components are briefly described next:

Variation operators are necessary to evolve the population. The main operators we use at the evolution stage are crossover and mutation. Extra operators can be easily added if necessary.

| **Algorithm:** Solving the Periodic Timetable Generation Problem |
|---|
| 1:  Generate a random population $P$ <br> 2:  **while** Termination condition is not met **do** <br> 3:      **for each** individual $i$ of $P$ **do** <br> 4:          Generate a timetable using individual $i$ <br> 5:          Evaluate the generated timetable using the fitness function <br> 6:      **end for** <br> 7:      Apply variation operators <br> 8:      Generate a new population $P$ <br> 9: **end while** |

**Algorithm 1. – Solving the Periodic Timetable Problem**

In order to calculate the fitness value of an individual, a timetable must be generated. For this, the values in the individual (vector) are used to calculate all arrival and departure times. After the timetable is generated, it is evaluated using a slightly simplified version of the objective function described in Section 3. Indeed running and dwelling constraints violations do not need to be evaluated since the individuals structure do not allow invalid values for both running and dwelling times.

The fitness function consists in minimizing the total weighted sum of constraints violations for the timetable.

## 5. EXPERIMENTAL ANALYSIS

For experimenting and validating our work, we propose two case studies. After describing them, a set of results is presented and discussed. The specifications of the computer used to solve both case studies are: Processor Intel Core i7 @2.9 GHz x4, 3.5 GB of RAM memory and running Ubuntu-64 v12.10 as operating system.

### 5.1. Description of the instances

#### 5.1.1. Case Study 1
We consider a subnetwork of the Netherlands rail system. This case study is obtained from Meester and Muns (2007). Mainly, it comprehends 4 train lines: 8 trains in total, 10 stations and the topology of the network is presented in Figure 3. As relevant information, the time period $T$ was established to 60 minutes, and there are 7 connection constraints, giving us a total of 65 constraints. The details regarding the line itineraries, their running, dwelling and connection times are detailed in the referenced paper.

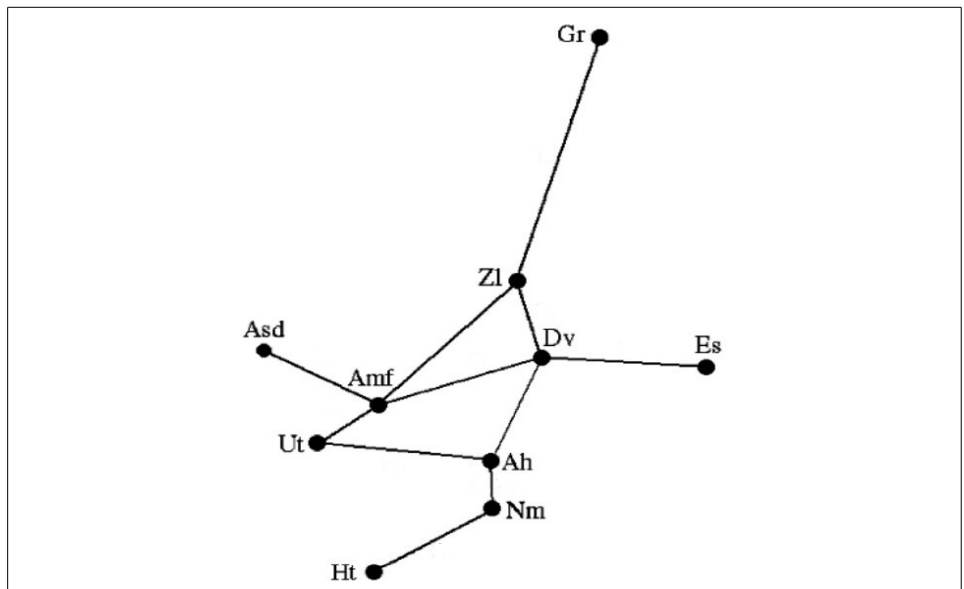

**Fig. 3 – Case Study 1: Subset of the Netherlands railway network**

#### 5.1.2. Case Study 2
It represents a simplified version of the *Nord-Pas-de-Calais* regional rail network, it means that only the main stations are taken into account (Figure 4). The data was collected from the different current timetables available to download from the SNCF website (2013). The time period $T$ was established to 60 minutes, the simplified network comprehends 26 stations and 24 train lines, i.e., a total of 48 trains to be scheduled. Additionally, 14 connection constraints are defined, and this gives us a total of 452 constraints.

## 5.2. Results

### 5.2.1. Case Study 1

For the experimentation, we use three sizes of population: 300, 600 and 900 individuals. Additionally, we set as stop-criteria, that either an optimal solution is found or an evaluation count limit is reached. The evaluation count represents the number of times a fitness value is calculated, in our case, how many timetables are generated. The evaluation count limits were set to: 10K, 20K, 30K, 200K, 1M and 5M. For each permutation of population with the evaluation count limits, the algorithm was executed 50 times, the results are shown in Table 4.

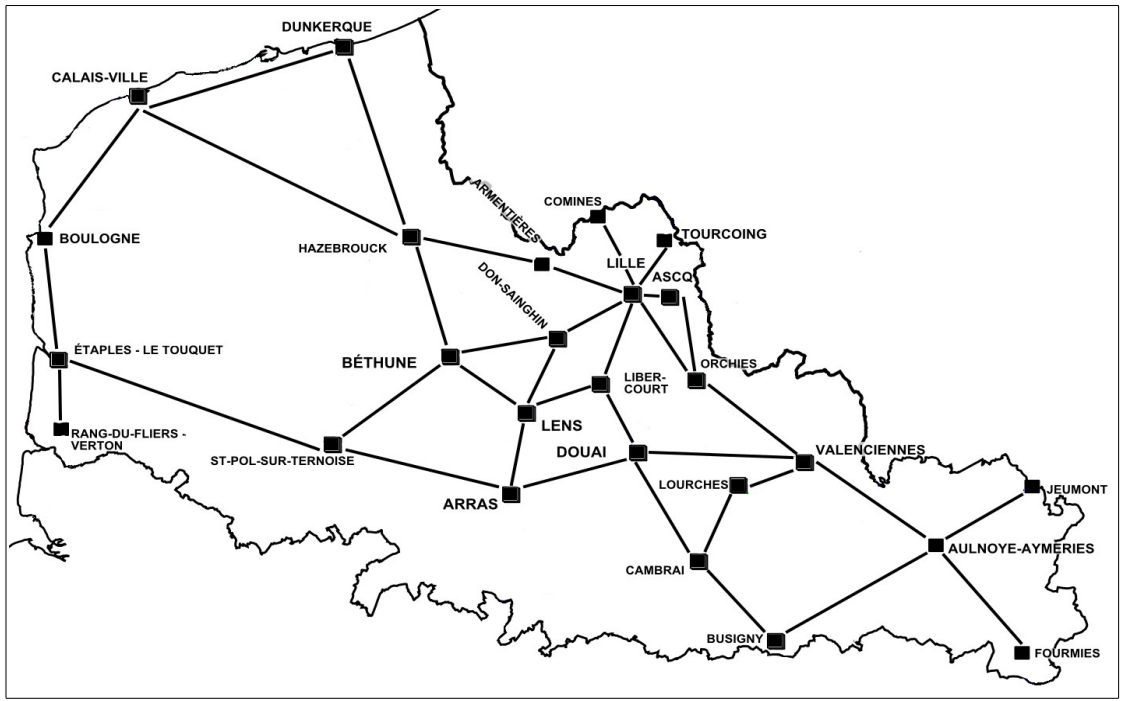

**Fig. 4 – Simplified Nord-Pas-de-Calais rail network**

Our solution can find a feasible timetable (without any hard constraint violation) in 100% of cases under 0.91 seconds. However, in average, 0.34 connection constraints are violated, which means that 66% of solutions generated under 0.91 seconds are feasible and observe also the connection constraints. Increasing the evaluation count limit allows our solution to generate better timetables: after 2.16 seconds, 90% of the generated timetables comply all constraints. It is around 7.26 seconds in average, that our solution is able to generate 100% of timetables that do not violate any constraint at all.

### 5.2.2. Case Study 2

For case study 2, the sizes of populations chosen were: 600, 800 and 1000, while the selected evaluation count limits were: 50K, 150K, 250K, 500K, 1M and 5M. Likewise the previous experimentation, each permutation of these two parameters was executed 50 times. The results are summarized in Table 5.

| Max Evaluations | Avg. Hard C. Violations | Avg. Soft C. Violations | Avg. Feasible (%) | Avg. Feasible + Connections (%) | Avg. Processing Time (s) |
|---|---|---|---|---|---|
| 10 K | 0 | 0.34 | 100 | 66 | 0.91 |
| 20 K | 0 | 0.25 | 100 | 75.33 | 1.19 |
| 30 K | 0 | 0.20 | 100 | 80 | 1.27 |
| 200 K | 0 | 0.10 | 100 | 90 | 2.16 |
| 1 M | 0 | 0.04 | 100 | 96 | 3.52 |
| 5 M | 0 | 0 | 100 | 100 | 7.26 |

**Table 4 – Results of experimentation, Case Study 1.**

As expected, this case study requires a larger number of evolutions in order to achieve an acceptable percentage of feasible timetables generated. 92.67% of the timetables generated after an average of 34.55 seconds were feasible, but only 46.67% of them respect all connection constraints. It is around 124.56 seconds in average that 99.33% of the generated timetables are not only feasible but also they observe all connection constraints. It is important to note that, even if the average elapsed time seems elevated, this is caused by those runs in which the solution must reach the limit of evaluations in order to stop. In many cases, when the optimal solution is found, i.e. zero constraints violated, the processing time is considerably lower than the average showed in the results.

| Max Evaluations | Avg. Hard C. Violations | Avg. Soft C. Violations | Avg. Feasible (%) | Avg. Feasible + Connections (%) | Avg. Processing Time (s) |
|---|---|---|---|---|---|
| 50 K | 0.47 | 2.98 | 58 | 1.33 | 16.77 |
| 150 K | 0.07 | 0.66 | 92.67 | 46.67 | 34.55 |
| 250 K | 0.05 | 0.47 | 95.33 | 58 | 46.88 |
| 500 K | 0.03 | 0.18 | 97.33 | 80 | 62.51 |
| 1 M | 0 | 0.11 | 100 | 88.67 | 77.84 |
| 5 M | 0 | 0.01 | 100 | 99.33 | 124.56 |

**Table 5 – Results of experimentation, Case Study 2.**

## 6. CONCLUSION AND PERSPECTIVES

The periodic train timetable design represents a complex research subject for which there are many different ways to address, model and solve the problem. The results of these may improve the current train timetable design system in the countries exploiting periodic timetables.

The results of our implementation to solve the periodic train timetable generation problem are very encouraging. Indeed, even though it represents a hard and complex problem, the performance of the genetic algorithm is very satisfactory. Our solution not only can rapidly find a feasible timetable for the problem, but it also offers a flexible platform in which we can easily add, remove or modify the constraints in order to satisfy some other criteria. For example, it can be easily turned into a solution that focuses on the reduction of the overall waiting time for passengers. Also, by modifying the weight values of the constraint violations, we can focus on the improvement of a particular type of constraints, for example by optimizing the connections between lines in the rail network.

However, our current solution does not deal with a very important issue in the train timetable design, which is the planning of resources such as the track layout inside the stations, the number of platforms or the available rolling stock. Thus, further improvements to our approach should extend our model allowing it to handle this kind of resources in order to offer a more realistic solution.

As stated before, we consider that the algorithm performance solving the timetabling problem was quite good, but in order to validate and compare it with other approaches, a benchmarking process is required. It is then, as a next step for us: first to find and use a benchmark platform used by other authors solving the same type problem, then to modify and adapt our solution to this benchmark in order to finally make a fair comparison of performance.

Additionally, we could improve our implementation according to a multi-objective paradigm, that is, deeply studying and analyzing how some constraints affect the final timetable. We could produce different kinds of options that would specialize in some particular interest, like optimization or finding a better connection plan to ensure the desired connectivity in the network. We believe that it can be done by adding new kinds of constraints or modifying the existing ones.

Finally, we are also considering to combine our genetic algorithm with other optimization technique, such as mathematical programming, to create a hybrid method. This can be made, not only to overcome the limitations of the genetic algorithms, e.g., a tendency to converge towards local optimal solutions, but also to take advantage of the strengths of both methods.